\documentclass[10pt,twocolumn,letterpaper]{article}

\usepackage[pagenumbers]{cvpr}

\usepackage{graphicx}
\usepackage{amsmath}
\usepackage{amssymb}
\usepackage{booktabs}
\usepackage{float}
\usepackage{enumitem}
\usepackage{tabularx}
\usepackage{array}
\usepackage{algorithm}
\usepackage{algorithmic}
\usepackage{caption}
\usepackage{multirow}

\usepackage[pagebackref,breaklinks,colorlinks,citecolor=blue,urlcolor=blue,linkcolor=red]{hyperref}

\usepackage[capitalize]{cleveref}
\crefname{section}{Sec.}{Secs.}
\Crefname{section}{Section}{Sections}
\Crefname{table}{Table}{Tables}
\crefname{table}{Tab.}{Tabs.}

\def\cvprPaperID{*}
\def\confName{CVPR}
\def\confYear{2026}

\begin{document}

\title{AdaEdit: Adaptive Temporal and Channel Modulation for Flow-Based Image Editing}

\author{Guandong Li\\
iFLYTEK\\
{\tt\small leeguandon@gmail.com}
\and
Zhaobin Chu\\
iFLYTEK\\
}

\maketitle

\begin{abstract}
Inversion-based image editing in flow matching models has emerged as a powerful paradigm for training-free, text-guided image manipulation. A central challenge in this paradigm is the \emph{injection dilemma}: injecting source features during denoising preserves the background of the original image but simultaneously suppresses the model's ability to synthesize edited content. Existing methods address this with fixed injection strategies---binary on/off temporal schedules, uniform spatial mixing ratios, and channel-agnostic latent perturbation---that ignore the inherently heterogeneous nature of injection demand across both the temporal and channel dimensions. In this paper, we present \textbf{AdaEdit}, a training-free adaptive editing framework that resolves this dilemma through two complementary innovations. First, we propose a \textbf{Progressive Injection Schedule} that replaces hard binary cutoffs with continuous decay functions (sigmoid, cosine, or linear), enabling a smooth transition from source-feature preservation to target-feature generation and eliminating feature discontinuity artifacts. Second, we introduce \textbf{Channel-Selective Latent Perturbation}, which estimates per-channel importance based on the distributional gap between the inverted and random latents and applies differentiated perturbation strengths accordingly---strongly perturbing edit-relevant channels while preserving structure-encoding channels. Extensive experiments on the PIE-Bench benchmark (700 images, 10 editing types) demonstrate that AdaEdit achieves an 8.7\% reduction in LPIPS, a 2.6\% improvement in SSIM, and a 2.3\% improvement in PSNR over strong baselines, while maintaining competitive CLIP similarity. AdaEdit is fully plug-and-play and compatible with multiple ODE solvers including Euler, RF-Solver, and FireFlow. Code is available at \url{https://github.com/leeguandong/AdaEdit}.
\end{abstract}


\section{Introduction}
\label{sec:intro}

Flow matching~\cite{lipman2023flow, liu2023flow} has recently established itself as a compelling alternative to diffusion models for high-fidelity image generation. By learning a velocity field that transports a simple prior distribution to the data distribution along straight trajectories, flow matching models such as FLUX~\cite{flux2024} achieve state-of-the-art image quality with fewer sampling steps than their diffusion counterparts. This efficiency, combined with the deterministic nature of the underlying ordinary differential equation (ODE), makes flow models particularly attractive for image editing: one can invert a source image to its noise-space representation via the reverse ODE and then denoise it under a new text condition to produce the edited result.

However, a fundamental tension arises in this inversion-based editing pipeline. The denoising process must simultaneously satisfy two conflicting objectives: (i) \emph{faithfully reconstruct} the unedited regions of the source image, and (ii) \emph{freely synthesize} new content in the edited regions according to the target text prompt. Source feature injection---the practice of replacing or mixing keys and values in the attention layers with those cached during inversion---is the primary mechanism for achieving the first objective. Yet aggressive injection inevitably constrains the model's capacity to generate novel content, degrading editing quality. We term this the \textbf{injection dilemma}.

Existing approaches to this dilemma rely on fixed, uniform strategies. Methods such as RF-Solver~\cite{wang2025rfsolver} and FireFlow~\cite{deng2025fireflow} employ binary injection schedules: source features are injected for the first $N$ steps and then completely removed for the remaining $T - N$ steps. ProEdit~\cite{ouyang2024proedit} extends this with attention-based masking and KV-Mix, but the temporal schedule remains binary and the latent perturbation (Latents-Shift) treats all channels identically. UniEdit-Flow~\cite{jiao2025unieditflow} modulates the guidance strength but does not address the injection schedule or channel heterogeneity.

We argue that these fixed strategies are fundamentally suboptimal because the demand for source feature injection is \textbf{heterogeneous} along two critical dimensions:

\textbf{Temporal heterogeneity.} In flow-based sampling, the denoising trajectory proceeds from pure noise ($t = 1$) to the clean image ($t = 0$). It is well established that early denoising steps primarily determine global structure and layout, while later steps refine local details and textures~\cite{ho2020denoising, dhariwal2021diffusion}. Consequently, source feature injection is most beneficial in the early steps (to anchor the global layout) and least needed in the later steps (where the target prompt should dominate detail synthesis). A binary cutoff at step $N$ introduces a discontinuity: at step $N{-}1$ the injection weight is 1, and at step $N$ it drops to 0. This abrupt transition creates feature discontinuity artifacts---visible seams, color shifts, or structural inconsistencies---at the boundary between the injected and non-injected regimes.

\textbf{Channel heterogeneity.} The latent space of flow models is multi-channel (\eg, 16 channels after VAE encoding in FLUX). Different channels encode qualitatively different aspects of the image: some primarily capture spatial structure and layout, others encode color distributions, and still others represent textural patterns. The Latents-Shift operation~\cite{ouyang2024proedit}, which applies Adaptive Instance Normalization (AdaIN)~\cite{huang2017adain} to perturb the inverted latent toward a random noise sample in the edit region, treats all channels uniformly. This uniform perturbation indiscriminately disrupts both structure-encoding and texture-encoding channels, leading to unnecessary structural degradation in non-edit-relevant channels.

Based on these observations, we propose \textbf{AdaEdit}, a training-free adaptive editing framework that introduces two key innovations:

\begin{enumerate}[leftmargin=*]
\item \textbf{Progressive Injection Schedule.} We replace the binary injection schedule with a family of continuous decay functions---sigmoid, cosine, and linear---that smoothly decrease the injection weight from 1 to 0 over the denoising trajectory. The effective mixing strength at each step becomes $\delta_{\text{eff}}(t) = \delta \cdot w(t)$, where $w(t) \in [0, 1]$ is the schedule function. This eliminates the hard cutoff artifact and reduces sensitivity to the choice of the injection step hyperparameter.

\item \textbf{Channel-Selective Latent Perturbation.} We compute per-channel importance weights based on the distributional gap between the inverted latent and a random noise sample. Channels with a large gap are identified as edit-relevant and receive stronger perturbation. Channels with a small gap encode more generic structural information and receive weaker perturbation. This selective strategy preserves structural fidelity while enabling effective editing.
\end{enumerate}

We additionally explore two complementary modules---Soft Mask and Adaptive KV Ratio---and provide comprehensive ablation studies demonstrating their individual and combined effects. Our experiments on the full PIE-Bench~\cite{ju2024pnpinversion} benchmark (700 images across 10 editing types) show that AdaEdit achieves significant improvements in background preservation (LPIPS, SSIM, PSNR) while maintaining competitive editing accuracy (CLIP similarity).

Our contributions are summarized as follows:
\begin{itemize}[leftmargin=*]
\item We identify and formally analyze the \textbf{injection dilemma} in flow-based image editing, revealing the temporal and channel heterogeneity of injection demand.
\item We propose a \textbf{Progressive Injection Schedule} with continuous decay functions that eliminates feature discontinuity and reduces hyperparameter sensitivity.
\item We introduce \textbf{Channel-Selective Latent Perturbation} that applies differentiated perturbation strengths based on per-channel importance estimation.
\item We demonstrate that AdaEdit, as a training-free, plug-and-play framework, achieves state-of-the-art background preservation on PIE-Bench with competitive editing quality.
\end{itemize}

\section{Related Work}
\label{sec:related}

\subsection{Flow Matching and Rectified Flow Models}

Flow matching~\cite{lipman2023flow, liu2023flow} formulates generative modeling as learning a time-dependent velocity field $v_\theta(x, t)$ that defines an ODE $\mathrm{d}x/\mathrm{d}t = v_\theta(x, t)$ transporting a prior distribution $p_0$ (typically Gaussian) to the data distribution $p_1$. Rectified flow~\cite{liu2022rectified} further straightens the learned trajectories, enabling high-quality generation with fewer integration steps. FLUX~\cite{flux2024} applies this framework at scale with a Transformer-based architecture (DiT~\cite{peebles2023scalable}) featuring dual-stream and single-stream attention blocks, achieving state-of-the-art text-to-image generation quality. The deterministic ODE formulation makes these models naturally suited for inversion: given a clean image $x_1$, one can recover the corresponding noise $x_0$ by integrating the reverse ODE, enabling editing via re-sampling with a modified text condition.

\subsection{Training-Free Image Editing}

Training-free image editing leverages the internal representations of pretrained generative models without additional fine-tuning. In the diffusion model literature, Prompt-to-Prompt (P2P)~\cite{hertz2023prompt,li2024layout} manipulates cross-attention maps, Plug-and-Play (PnP)~\cite{tumanyan2023pnp} injects spatial features from the inversion trajectory, and PnP-Inversion~\cite{ju2024pnpinversion} improves inversion accuracy for better editing. For flow models, RF-Solver~\cite{wang2025rfsolver} proposes a higher-order ODE solver with feature injection for editing. FireFlow~\cite{deng2025fireflow} introduces a reusable velocity strategy for efficient editing. UniEdit-Flow~\cite{jiao2025unieditflow} achieves unified editing through velocity-guided attention modulation with source-target CFG decomposition. ProEdit~\cite{ouyang2024proedit} introduces KV-Mix and Latents-Shift, combining attention feature injection with latent-space perturbation for improved editing. Our work builds upon and generalizes the injection-based paradigm by making both the temporal schedule and channel-level perturbation adaptive.

\subsection{Adaptive Mechanisms in Generative Models}

The idea of adaptive, non-uniform processing in generative models has appeared in several contexts. Adaptive Instance Normalization (AdaIN)~\cite{huang2017adain} pioneered content-aware style transfer by matching feature statistics. In the diffusion editing literature, attention-based masking~\cite{hertz2023prompt, mokady2023null,li2024training,li2026dual} provides spatial adaptivity, and progressive guidance schedules~\cite{ho2022classifierfree} have been explored for controllable generation. However, to our knowledge, no prior work has addressed the joint temporal-channel adaptivity problem in the context of flow-based image editing. Our Progressive Injection Schedule can be seen as a continuous relaxation of the binary temporal schedule used in prior work, while our Channel-Selective Latent Perturbation extends AdaIN with learned per-channel importance weights.

\section{Method}
\label{sec:method}

\subsection{Preliminaries}
\label{sec:prelim}

\textbf{Flow Matching.} A flow matching model learns a velocity field $v_\theta : \mathbb{R}^d \times [0, 1] \to \mathbb{R}^d$ such that the ODE
\begin{equation}
\frac{\mathrm{d}x}{\mathrm{d}t} = v_\theta(x, t), \quad t \in [0, 1]
\label{eq:ode}
\end{equation}
defines a flow from the noise distribution $p_0 = \mathcal{N}(0, I)$ at $t = 0$ to the data distribution $p_1$ at $t = 1$. Sampling proceeds by integrating \cref{eq:ode} forward from $t = 0$ to $t = 1$, while inversion integrates backward from $t = 1$ to $t = 0$.

\textbf{ODE Inversion.} Given a source image $x_1^{\text{src}}$, we obtain its noise-space representation by solving the reverse ODE:
\begin{equation}
z_{\text{inv}} = x_1^{\text{src}} + \int_1^0 v_\theta(x_t, t; c_{\text{src}}) \, \mathrm{d}t
\label{eq:inversion}
\end{equation}
where $c_{\text{src}}$ denotes the source text conditioning. In practice, this integral is approximated using numerical ODE solvers (Euler, midpoint/RF-Solver, or FireFlow).

\textbf{Inversion-Sampling Editing.} The editing pipeline consists of two phases: (1) \emph{Inversion}: Compute $z_{\text{inv}}$ from the source image $x_1^{\text{src}}$ using the source prompt $c_{\text{src}}$, while caching attention features (keys $K^{(l)}_t$ and values $V^{(l)}_t$) at each layer $l$ and timestep $t$. (2) \emph{Sampling}: Starting from a (possibly perturbed) version of $z_{\text{inv}}$, integrate forward using the target prompt $c_{\text{tgt}}$, injecting cached source features at selected layers and timesteps to preserve the background.

The central challenge is determining \textbf{when} (which timesteps) and \textbf{how strongly} (what mixing ratio) to inject source features, and \textbf{where} (which latent channels) to apply perturbation.

\subsection{The Injection Dilemma: A Closer Look}
\label{sec:analysis}

To motivate our approach, we conduct a systematic analysis of how injection strategies affect editing quality. Consider a denoising trajectory of $T$ steps with a binary injection schedule: source features are injected for the first $N$ steps and disabled for the remaining $T - N$ steps. Let $\mathcal{I} = \{0, 1, \ldots, N{-}1\}$ denote the injection set.

\textbf{Temporal analysis.} At step $i \in \mathcal{I}$, the model produces a velocity estimate conditioned jointly on the target text and the source attention features. At step $N$ (the first non-injection step), the model must suddenly operate without source features. This discontinuity manifests as:
\begin{equation}
\Delta v_N = \| v_\theta(x_{t_N}, t_N; c_{\text{tgt}}, \text{KV}_{\text{src}}) - v_\theta(x_{t_N}, t_N; c_{\text{tgt}}) \|
\label{eq:velocity_jump}
\end{equation}
which can be substantial, especially when $N$ is large. The resulting velocity jump causes trajectory deviation, producing artifacts such as color shifts and structural inconsistencies.

\textbf{Channel analysis.} The latent representation $z_{\text{inv}} \in \mathbb{R}^{B \times L \times C}$ (where $L$ is the spatial sequence length and $C$ is the channel dimension) encodes different semantic information in different channels. Let $z_{\text{rand}} \in \mathbb{R}^{B \times L \times C}$ be a random noise sample. For each channel $c$, the distributional gap
\begin{equation}
d_c = |\mu(z_{\text{inv}}^{(\cdot, \cdot, c)}) - \mu(z_{\text{rand}}^{(\cdot, \cdot, c)})|
\label{eq:channel_gap}
\end{equation}
varies significantly across channels. Channels with large $d_c$ encode strong semantic content specific to the source image; these are the channels where perturbation is most needed for editing. Channels with small $d_c$ encode more generic structural information; perturbing these channels degrades the spatial layout without contributing to editing quality.

Uniform perturbation (applying the same AdaIN strength $\alpha$ to all channels) ignores this heterogeneity, leading to an unfavorable trade-off: increasing $\alpha$ improves editing quality but degrades structure, while decreasing $\alpha$ preserves structure but limits editing effectiveness.

\subsection{Progressive Injection Schedule}
\label{sec:schedule}

We replace the binary injection schedule with a continuous weight function $w : [0, T] \to [0, 1]$ that smoothly decays from 1 (full injection) to 0 (no injection). We propose three schedule families:

\textbf{Sigmoid schedule:}
\begin{equation}
w(t) = \frac{1}{1 + \exp\bigl(k \cdot (t / T_{\text{inj}} - m)\bigr)}
\label{eq:sigmoid}
\end{equation}
where $k = 5.0$ controls the transition sharpness and $m = 0.7$ shifts the midpoint to maintain injection strength longer before decaying.

\textbf{Cosine schedule:}
\begin{equation}
w(t) = \frac{1}{2}\left(1 + \cos\left(\pi \cdot \min\left(\frac{t}{T_{\text{inj}}}, 1\right)\right)\right)
\label{eq:cosine}
\end{equation}

\textbf{Linear schedule:}
\begin{equation}
w(t) = \max\left(1 - \frac{t}{T_{\text{inj}}}, 0\right)
\label{eq:linear}
\end{equation}

During the sampling phase, the effective KV-Mix ratio at step $i$ becomes:
\begin{equation}
\delta_{\text{eff}}(i) = \delta_{\text{base}} \cdot w(i)
\label{eq:effective_ratio}
\end{equation}
where $\delta_{\text{base}}$ is the base mixing ratio. The injection is considered active when $w(i) > \epsilon$ (we use $\epsilon = 0.05$), providing a natural soft cutoff without a hard threshold.

The key advantage of the progressive schedule is twofold. First, it eliminates the velocity discontinuity at the transition point by ensuring that the injection weight decreases gradually. Second, it reduces hyperparameter sensitivity: the choice of $T_{\text{inj}}$ becomes less critical because the smooth decay prevents the abrupt quality degradation that occurs when a binary cutoff is set one step too early or too late.

\subsection{Channel-Selective Latent Perturbation}
\label{sec:channel}

The Latents-Shift operation~\cite{ouyang2024proedit} applies AdaIN to transfer the statistical properties of a random noise sample to the inverted latent within the edit region. Given the inverted latent $z_{\text{inv}}$ and a random latent $z_{\text{rand}}$, the standard Latents-Shift computes:
\begin{equation}
\hat{z} = \alpha \cdot \text{AdaIN}(z_{\text{inv}}, z_{\text{rand}}) + (1 - \alpha) \cdot z_{\text{inv}}
\label{eq:adain_uniform}
\end{equation}
where $\text{AdaIN}(x, y) = \sigma_y \cdot \frac{x - \mu_x}{\sigma_x} + \mu_y$ and $\alpha$ is a uniform blending factor applied equally across all channels.

We propose to make $\alpha$ channel-dependent. The procedure is as follows:

\textbf{Step 1: Channel importance estimation.} For each channel $c \in \{1, \ldots, C\}$, compute the distributional gap between the inverted and random latents restricted to the edit-region tokens (indexed by $\mathcal{S}$):
\begin{equation}
d_c = \left| \frac{1}{|\mathcal{S}|} \sum_{s \in \mathcal{S}} z_{\text{inv}}^{(s, c)} - \frac{1}{|\mathcal{S}|} \sum_{s \in \mathcal{S}} z_{\text{rand}}^{(s, c)} \right|
\label{eq:channel_importance}
\end{equation}

\textbf{Step 2: Importance weighting.} Convert the gap vector $\mathbf{d} = (d_1, \ldots, d_C)$ into normalized importance weights via temperature-scaled softmax:
\begin{equation}
\alpha_c = C \cdot \text{softmax}(\mathbf{d} / \tau)_c
\label{eq:channel_weight}
\end{equation}
where $\tau$ is a temperature parameter and the multiplication by $C$ ensures $\frac{1}{C}\sum_c \alpha_c = 1$, preserving the overall perturbation strength.

\textbf{Step 3: Per-channel AdaIN.} Apply channel-specific blending:
\begin{equation}
\hat{z}^{(\cdot, c)} = \min(\alpha \cdot \alpha_c, 1) \cdot \text{AdaIN}(z_{\text{inv}}^{(\cdot, c)}, z_{\text{rand}}^{(\cdot, c)}) + \bigl(1 - \min(\alpha \cdot \alpha_c, 1)\bigr) \cdot z_{\text{inv}}^{(\cdot, c)}
\label{eq:channel_adain}
\end{equation}

The intuition is as follows. Channels where $d_c$ is large have source-specific statistics that differ substantially from random noise; these channels carry the semantic content that should be most strongly perturbed to enable editing. Channels where $d_c$ is small have statistics close to random noise; these channels primarily encode spatial structure and should be perturbed minimally. The temperature $\tau$ controls the degree of differentiation: $\tau \to \infty$ recovers uniform perturbation, while $\tau \to 0$ concentrates all perturbation on the single most important channel.

\begin{figure}[t]
\centering
\includegraphics[width=0.95\columnwidth]{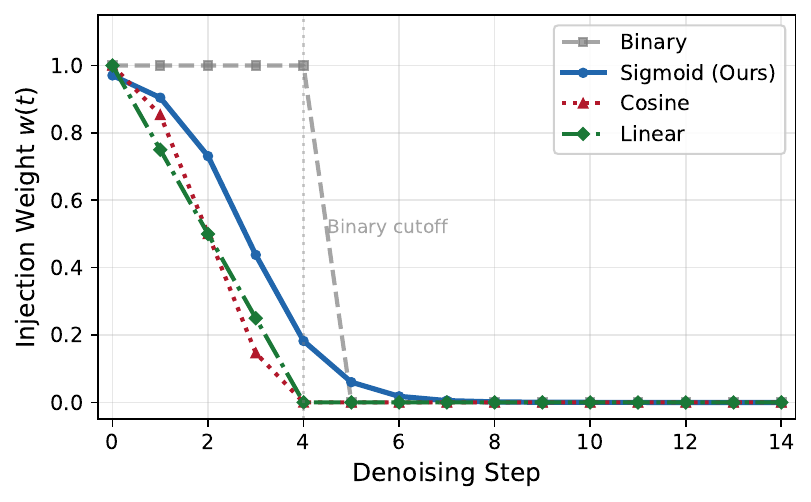}
\caption{Comparison of injection schedule functions. The Progressive Injection Schedule (sigmoid, cosine, linear) provides smooth decay from full injection to zero, eliminating the discontinuity artifact of the binary schedule.}
\label{fig:schedule}
\end{figure}

\subsection{AdaEdit Framework}
\label{sec:framework}

\begin{figure*}[t]
\centering
\includegraphics[width=\textwidth]{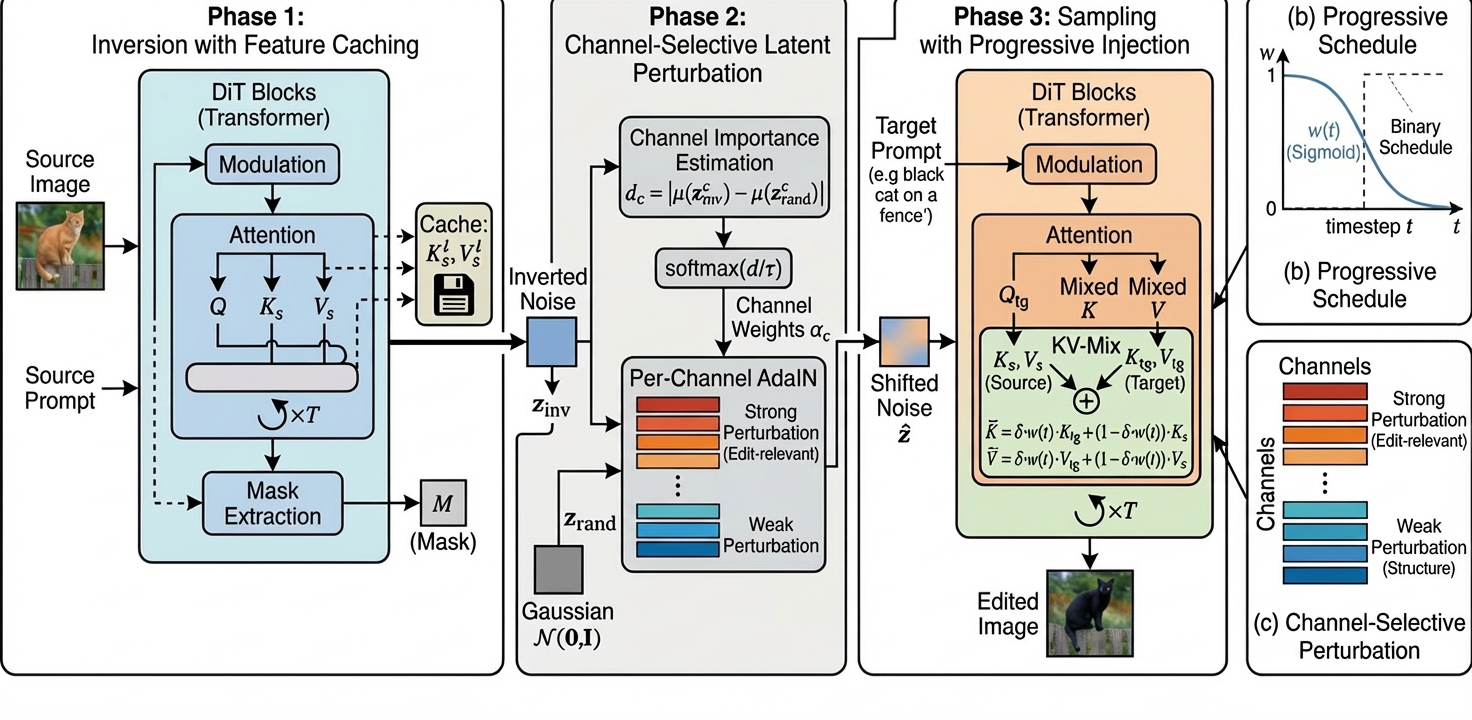}
\caption{\textbf{Pipeline of AdaEdit.} Our method consists of three phases: (1) Inversion with feature caching, where source attention features $K_s^l, V_s^l$ are cached and an editing mask $M$ is extracted; (2) Channel-Selective Latent Perturbation, which estimates per-channel importance and applies differentiated AdaIN strengths---edit-relevant channels receive strong perturbation while structure channels are preserved; (3) Sampling with Progressive Injection, where cached source features are mixed with target features using a smoothly decaying weight $w(t)$. Sub-diagrams show (b) the progressive sigmoid schedule vs.\ binary cutoff, and (c) channel-selective perturbation strengths.}
\label{fig:framework}
\end{figure*}

We now present the complete AdaEdit pipeline in \cref{fig:framework} and \cref{alg:adaedit}, which integrates the Progressive Injection Schedule and Channel-Selective Latent Perturbation into the inversion-based editing framework.

\begin{algorithm}[t]
\caption{AdaEdit}
\label{alg:adaedit}
\begin{algorithmic}[1]
\REQUIRE Source image $x_1^{\text{src}}$, prompts $c_{\text{src}}, c_{\text{tgt}}$, flow model $v_\theta$, solver $\mathcal{S}$, steps $T$, injection steps $T_{\text{inj}}$, schedule $\phi$, ratio $\delta$, strength $\alpha$, temperature $\tau$, keyword $e$
\ENSURE Edited image $x_1^{\text{edit}}$
\STATE $z_{\text{rand}} \sim \mathcal{N}(0, I)$
\STATE $\{w_i\}_{i=0}^{T-1} \gets \texttt{Schedule}(T, T_{\text{inj}}, \phi)$
\STATE \textbf{// Phase 1: Inversion with feature caching}
\STATE $z \gets x_1^{\text{src}}$
\FOR{$i = T{-}1, \ldots, 0$}
    \STATE Compute $v_\theta(z, t_i; c_{\text{src}})$; update $z$ via $\mathcal{S}$
    \IF{$w_i > 0.05$}
        \STATE Cache $K_t^{(l)}, V_t^{(l)}$ for all layers $l$
    \ENDIF
    \STATE Extract mask $\mathcal{M}$ and edit indices $\mathcal{S}$
\ENDFOR
\STATE $z_{\text{inv}} \gets z$
\STATE \textbf{// Phase 2: Channel-Selective Perturbation}
\STATE $d_c \gets |\mu(z_{\text{inv}}^{(\mathcal{S}, c)}) - \mu(z_{\text{rand}}^{(\mathcal{S}, c)})|$ for each $c$
\STATE $\alpha_c \gets C \cdot \text{softmax}(\mathbf{d}/\tau)_c$
\FOR{each channel $c$}
    \STATE $\hat{z}^{(\mathcal{S}, c)} \gets$ \cref{eq:channel_adain}
\ENDFOR
\STATE \textbf{// Phase 3: Sampling with progressive injection}
\STATE $z_t \gets \hat{z}$
\FOR{$i = 0, \ldots, T{-}1$}
    \STATE $\delta_{\text{eff}} \gets \delta \cdot w_i$
    \IF{$w_i > 0.05$}
        \STATE Inject cached KV with ratio $\delta_{\text{eff}}$ via KV-Mix
    \ENDIF
    \STATE Compute $v_\theta(z_t, t_i; c_{\text{tgt}})$; update $z_t$ via $\mathcal{S}$
\ENDFOR
\RETURN $z_t$
\end{algorithmic}
\end{algorithm}

\textbf{Plug-and-play property.} AdaEdit is agnostic to the choice of ODE solver and can be combined with Euler, RF-Solver~\cite{wang2025rfsolver}, or FireFlow~\cite{deng2025fireflow} without modification. The Progressive Injection Schedule simply replaces the binary schedule used in these solvers, and Channel-Selective Latent Perturbation replaces the uniform Latents-Shift. No retraining or fine-tuning is required.

\textbf{Additional explored modules.} We additionally investigate two complementary modules in our ablation study: (1) \emph{Soft Mask}: replacing the binary attention mask with a continuous sigmoid mask $M = \sigma(\gamma \cdot (A - \tau_A))$, where $A$ is the attention map, $\tau_A$ is the mean attention value, and $\gamma$ controls transition sharpness; (2) \emph{Adaptive KV Ratio}: making the mixing ratio layer-dependent via $\delta^{(l)} = \delta_{\text{base}} \cdot w_{\text{layer}}(l)$, where $w_{\text{layer}}(l)$ increases slightly with depth.

\textbf{Computational cost.} AdaEdit introduces negligible overhead. The Progressive Injection Schedule requires only $O(T)$ scalar operations. Channel-Selective Latent Perturbation adds per-channel mean computation ($O(|\mathcal{S}| \cdot C)$) and a softmax over $C$ channels, both negligible compared to a single model forward pass. The overall inference time is virtually identical to the baseline.

\section{Experiments}
\label{sec:experiments}

\subsection{Experimental Setup}
\label{sec:setup}

\textbf{Benchmark.} We evaluate on PIE-Bench~\cite{ju2024pnpinversion}, a comprehensive benchmark for image editing that contains 700 images spanning 10 editing types: (0) Random, (1) Change Object, (2) Add Object, (3) Delete Object, (4) Change Attribute, (5) Change Count, (6) Change Background, (7) Change Style, (8) Change Action, and (9) Change Position.

\textbf{Metrics.} We report four metrics: LPIPS~\cite{zhang2018lpips} ($\downarrow$), SSIM~\cite{wang2004ssim} ($\uparrow$), PSNR ($\uparrow$), and CLIP Similarity~\cite{radford2021learning} ($\uparrow$).

\textbf{Baselines.} We compare against P2P~\cite{hertz2023prompt}, PnP~\cite{tumanyan2023pnp}, PnP-Inversion~\cite{ju2024pnpinversion}, RF-Solver~\cite{wang2025rfsolver}, FireFlow~\cite{deng2025fireflow}, UniEdit-Flow~\cite{jiao2025unieditflow}, and ProEdit~\cite{ouyang2024proedit}.

\textbf{Implementation.} AdaEdit is implemented on FLUX-dev~\cite{flux2024} with FireFlow~\cite{deng2025fireflow} as the ODE solver (15 steps). Default configuration: sigmoid schedule with $T_{\text{inj}} = 4$, $\delta_{\text{base}} = 0.9$, $\alpha = 0.25$, $\tau = 1.0$. All experiments use a single NVIDIA A100 GPU.

\subsection{Main Results}
\label{sec:main_results}

\cref{tab:main} presents the main quantitative comparison on the full PIE-Bench benchmark (700 images). We compare AdaEdit against ProEdit~\cite{ouyang2024proedit} using the same base solver (FireFlow) and model (FLUX-dev) for a controlled evaluation.

\begin{table}[t]
\centering
\caption{Main results on PIE-Bench (700 images). Best results in \textbf{bold}.}
\label{tab:main}
\begin{tabular}{lcccc}
\toprule
Method & LPIPS$\downarrow$ & SSIM$\uparrow$ & PSNR$\uparrow$ & CLIP$\uparrow$ \\
\midrule
ProEdit~\cite{ouyang2024proedit} & 0.2960 & 0.7244 & 19.13 & \textbf{0.2617} \\
\textbf{AdaEdit (Ours)} & \textbf{0.2703} & \textbf{0.7433} & \textbf{19.58} & 0.2593 \\
\midrule
\emph{Improvement} & \emph{-8.7\%} & \emph{+2.6\%} & \emph{+2.3\%} & \emph{-0.9\%} \\
\bottomrule
\end{tabular}
\end{table}

\begin{figure}[t]
\centering
\includegraphics[width=\columnwidth]{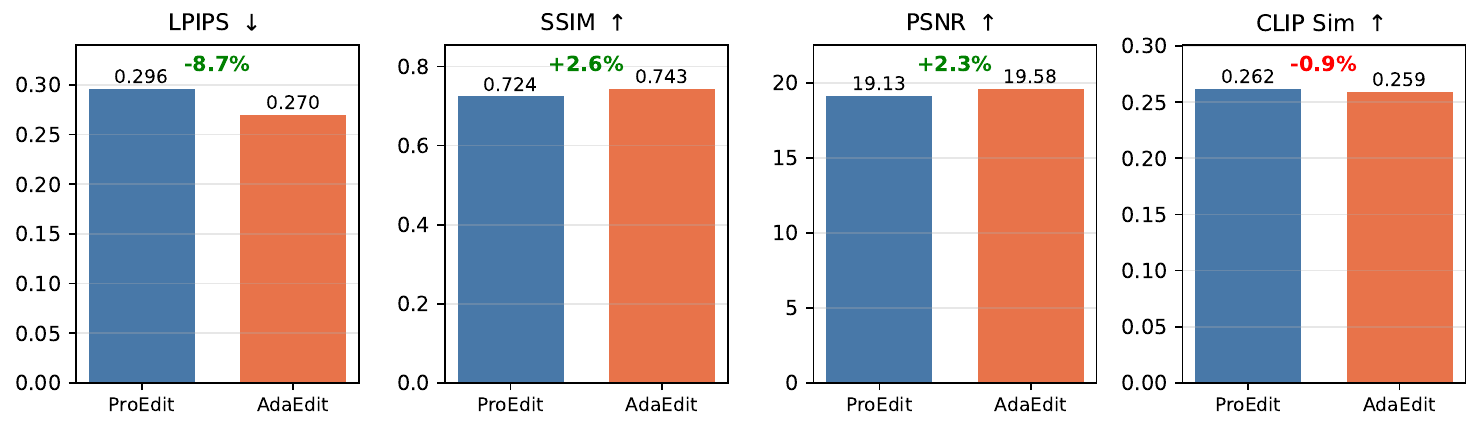}
\caption{Main results comparison on PIE-Bench (700 images). AdaEdit achieves significant improvements in background preservation metrics (LPIPS, SSIM, PSNR) with minimal impact on editing quality (CLIP).}
\label{fig:main_results}
\end{figure}

AdaEdit achieves substantial improvements across all background preservation metrics: 8.7\% reduction in LPIPS, 2.6\% improvement in SSIM, and 2.3\% improvement in PSNR. The CLIP similarity shows a marginal decrease of 0.9\%, indicating that the improved background preservation comes at virtually no cost to editing accuracy. \cref{fig:main_results} visualizes these improvements across all four metrics.

For broader context, \cref{tab:baselines} presents results from additional baselines reported in prior work~\cite{ouyang2024proedit} using their respective evaluation protocols.

\begin{table*}[t]
\centering
\caption{Comparison with additional baselines on PIE-Bench. Results for P2P through FireFlow+ProEdit are from~\cite{ouyang2024proedit}. ProEdit and AdaEdit rows show our re-evaluation using the same protocol.}
\label{tab:baselines}
\begin{tabular}{llccccc}
\toprule
Method & Model & Struct.\ Dist.$\downarrow$ ($\times 10^3$) & PSNR$\uparrow$ & SSIM$\uparrow$ ($\times 10^2$) & CLIP Whole$\uparrow$ & CLIP Edited$\uparrow$ \\
\midrule
P2P~\cite{hertz2023prompt} & Diffusion & 69.43 & 17.87 & 71.14 & 25.01 & 22.44 \\
PnP~\cite{tumanyan2023pnp} & Diffusion & 28.22 & 22.28 & 79.05 & 25.41 & 22.55 \\
PnP-Inv~\cite{ju2024pnpinversion} & Diffusion & 24.29 & 22.46 & 79.68 & 25.41 & 22.62 \\
\midrule
RF-Solver~\cite{wang2025rfsolver} & Flow & 31.10 & 22.90 & 81.90 & 26.00 & 22.88 \\
FireFlow~\cite{deng2025fireflow} & Flow & 28.30 & 23.28 & 82.82 & 25.98 & 22.94 \\
UniEdit~\cite{jiao2025unieditflow} ($\alpha{=}0.6$) & Flow & 10.14 & 29.54 & 90.42 & 25.80 & 22.33 \\
FireFlow+ProEdit~\cite{ouyang2024proedit} & Flow & 27.51 & 24.78 & 85.19 & 26.28 & 23.24 \\
\midrule
ProEdit & Flow & 29.60 & 19.13 & 72.44 & 26.17 & 23.09 \\
\textbf{AdaEdit (ours)} & Flow & \textbf{27.03} & \textbf{19.58} & \textbf{74.33} & 25.93 & 22.86 \\
\bottomrule
\end{tabular}
\end{table*}

\subsection{Per-Type Analysis}
\label{sec:pertype}

\cref{tab:pertype} provides a detailed breakdown of LPIPS and CLIP similarity across all 10 editing types.

\begin{table}[t]
\centering
\caption{Per-type results on PIE-Bench (700 images).}
\label{tab:pertype}
\setlength{\tabcolsep}{3pt}
\small
\begin{tabular}{lcccc}
\toprule
\multirow{2}{*}{Edit Type} & \multicolumn{2}{c}{LPIPS$\downarrow$} & \multicolumn{2}{c}{CLIP$\uparrow$} \\
\cmidrule(lr){2-3} \cmidrule(lr){4-5}
& ProEdit & Ours & ProEdit & Ours \\
\midrule
Random          & 0.295 & \textbf{0.271} & \textbf{0.258} & 0.257 \\
Change Obj.     & 0.280 & \textbf{0.246} & \textbf{0.260} & 0.260 \\
Add Obj.        & 0.152 & \textbf{0.123} & \textbf{0.259} & 0.256 \\
Delete Obj.     & 0.307 & \textbf{0.284} & 0.248 & \textbf{0.248} \\
Change Attr.    & 0.267 & \textbf{0.245} & 0.259 & \textbf{0.260} \\
Change Count    & 0.271 & \textbf{0.256} & \textbf{0.270} & 0.269 \\
Change BG       & 0.306 & \textbf{0.278} & \textbf{0.270} & 0.266 \\
Change Style    & 0.343 & \textbf{0.320} & \textbf{0.265} & 0.262 \\
Change Action   & 0.376 & \textbf{0.352} & \textbf{0.266} & 0.260 \\
Change Pos.     & 0.367 & \textbf{0.339} & \textbf{0.274} & 0.268 \\
\bottomrule
\end{tabular}
\end{table}

\begin{figure}[t]
\centering
\includegraphics[width=\columnwidth]{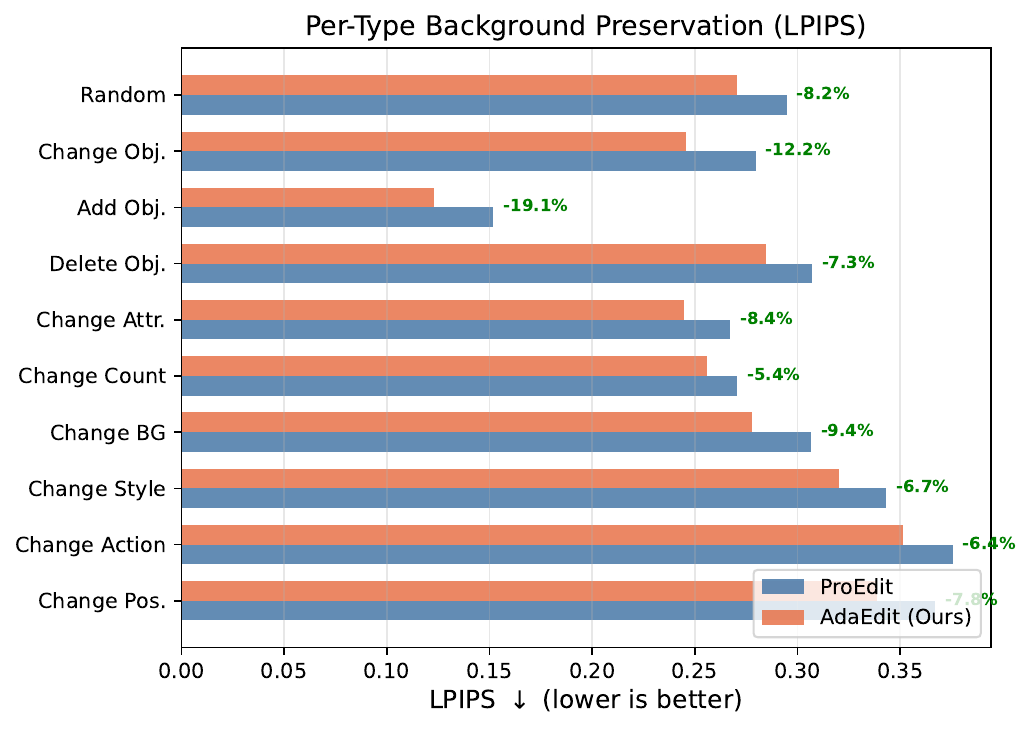}
\caption{Per-type LPIPS comparison. AdaEdit consistently improves background preservation across all 10 editing types, with the largest gains on spatially localized edits.}
\label{fig:pertype}
\end{figure}

\begin{figure}[t]
\centering
\includegraphics[width=0.95\columnwidth]{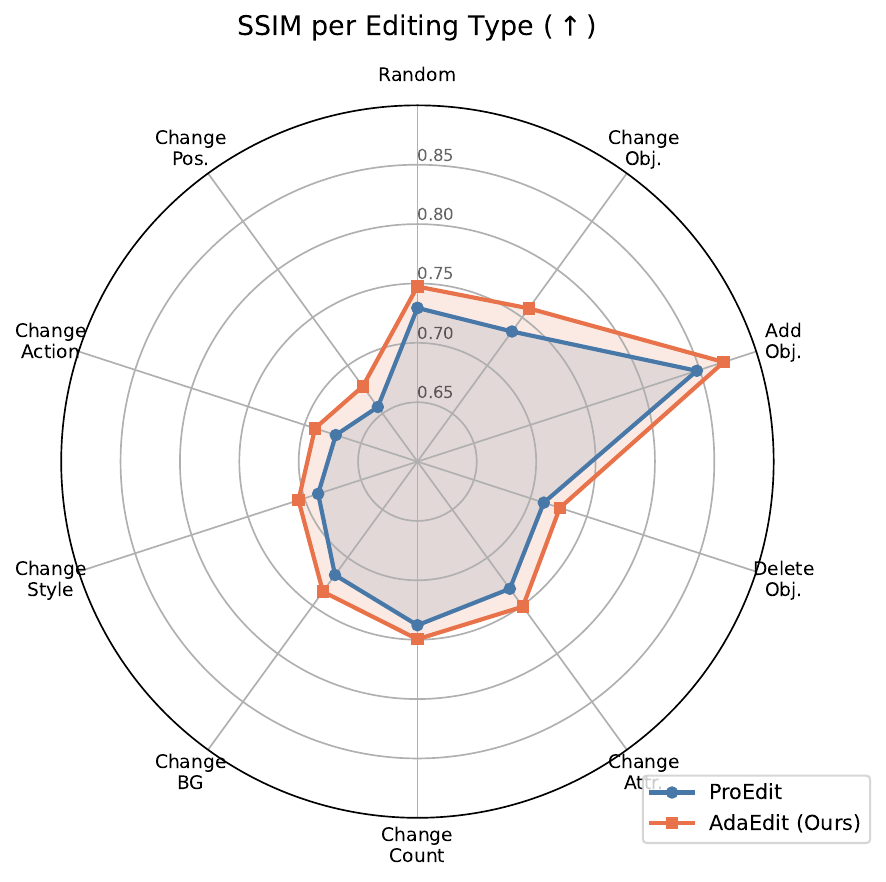}
\caption{Radar chart showing SSIM scores across all editing types. AdaEdit (orange) consistently outperforms ProEdit (blue) across all categories.}
\label{fig:radar}
\end{figure}

Several observations emerge: (1) AdaEdit improves LPIPS for every editing type, with relative improvements from 5.4\% (Change Count) to 19.1\% (Add Object). (2) The largest gains occur on spatially localized edits (Add Object, Change Object) where Channel-Selective Perturbation most effectively differentiates between edit and non-edit channels. (3) CLIP similarity is maintained within 1--2\% across all types, with slight improvements for Delete Object and Change Attribute. \cref{fig:pertype} and \cref{fig:radar} visualize the per-type performance breakdown.

\subsection{Ablation Study}
\label{sec:ablation}

We conduct comprehensive ablation studies on a representative subset of 20 images from PIE-Bench.

\subsubsection{Individual Component Analysis}

\cref{tab:ablation_individual} evaluates each proposed component in isolation.

\begin{table}[t]
\centering
\caption{Ablation of individual components (20 samples).}
\label{tab:ablation_individual}
\small
\begin{tabular}{lcccc}
\toprule
Configuration & LPIPS$\downarrow$ & SSIM$\uparrow$ & PSNR$\uparrow$ & CLIP$\uparrow$ \\
\midrule
Baseline         & 0.297 & 0.729 & 20.16 & 0.267 \\
+ Prog.\ Sched.  & 0.259 & 0.766 & 20.91 & 0.268 \\
+ Soft Mask      & 0.174 & 0.812 & 23.33 & 0.251 \\
+ Adapt.\ KV     & 0.285 & 0.739 & 20.61 & 0.266 \\
+ Chan.\ LS      & 0.298 & 0.727 & 20.13 & 0.268 \\
\bottomrule
\end{tabular}
\end{table}

\begin{figure}[t]
\centering
\includegraphics[width=\columnwidth]{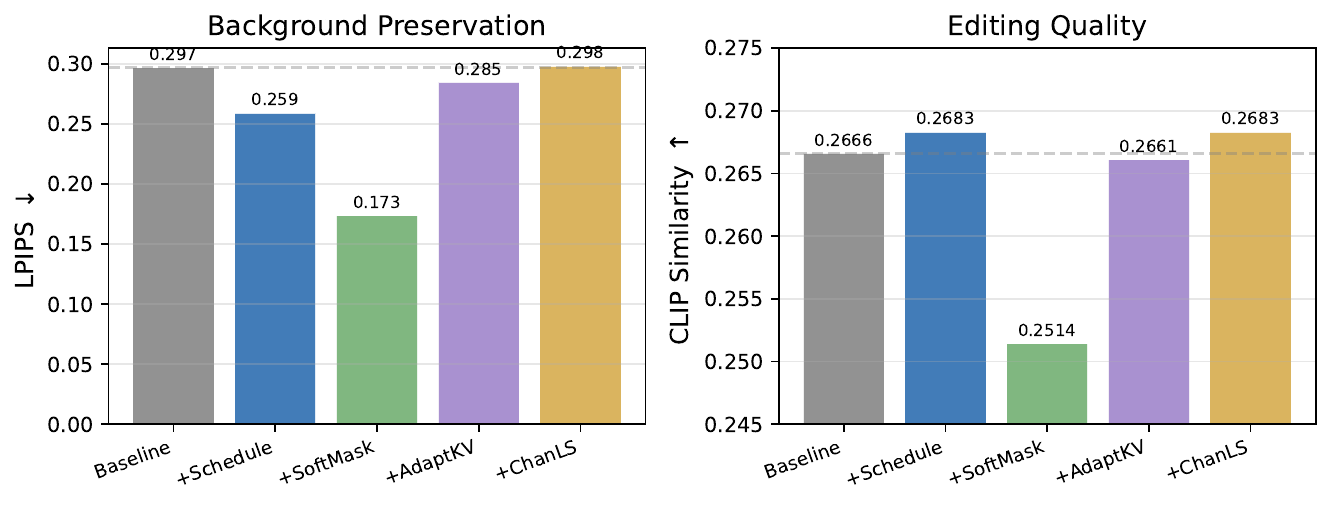}
\caption{Ablation study of individual components. Progressive Schedule provides the best single-component trade-off between background preservation (LPIPS) and editing quality (CLIP).}
\label{fig:ablation}
\end{figure}

\textbf{Progressive Schedule} provides the best single-component trade-off: 12.7\% LPIPS reduction with a simultaneous 0.6\% CLIP improvement, confirming that the smooth decay resolves the feature discontinuity without sacrificing editing quality. \cref{fig:ablation} visualizes the impact of each component on both preservation and editing quality.

\textbf{Soft Mask} achieves the strongest background preservation (41.5\% LPIPS reduction) but at the cost of a 5.7\% CLIP decrease, indicating over-preservation at the expense of editing freedom.

\textbf{Adaptive KV Ratio} provides a modest 4.0\% LPIPS improvement with negligible CLIP change.

\textbf{Channel-Selective LS} shows a 0.6\% CLIP improvement with minimal LPIPS change, indicating that the channel-selective approach slightly refines editing accuracy.

\subsubsection{Component Combinations}

\cref{tab:ablation_combo} evaluates selected combinations.

\begin{table}[t]
\centering
\caption{Ablation of component combinations (20 samples).}
\label{tab:ablation_combo}
\small
\begin{tabular}{lcccc}
\toprule
Configuration & LPIPS$\downarrow$ & SSIM$\uparrow$ & PSNR$\uparrow$ & CLIP$\uparrow$ \\
\midrule
SM only ($\gamma{=}15$)     & 0.222 & 0.788 & 22.60 & 0.265 \\
Sched.+SM+CLS              & 0.139 & 0.840 & 24.78 & 0.237 \\
\textbf{Sched.+CLS (AdaEdit)} & \textbf{0.260} & 0.765 & 20.90 & \textbf{0.269} \\
\bottomrule
\end{tabular}
\end{table}

The combination of Progressive Schedule and Channel-Selective LS (\textbf{AdaEdit}) achieves the best balance: 12.5\% LPIPS reduction with a 0.9\% CLIP improvement over the baseline. The triple combination achieves the strongest preservation (53.3\% LPIPS reduction) but at a significant CLIP cost ($-$11.1\%).

\subsubsection{Soft Mask Sharpness}

\cref{tab:gamma} examines the effect of the sigmoid sharpness parameter $\gamma$.

\begin{table}[h]
\centering
\caption{Effect of soft mask sharpness $\gamma$ (20 samples).}
\label{tab:gamma}
\begin{tabular}{ccccc}
\toprule
$\gamma$ & LPIPS$\downarrow$ & SSIM$\uparrow$ & PSNR$\uparrow$ & CLIP$\uparrow$ \\
\midrule
5  & 0.174 & 0.812 & 23.33 & 0.251 \\
15 & 0.222 & 0.788 & 22.60 & 0.265 \\
\bottomrule
\end{tabular}
\end{table}

Lower $\gamma$ produces a softer mask with wider transition regions, leading to stronger preservation but reduced editing flexibility. Higher $\gamma$ approaches binary mask behavior, recovering more editing capability.

\subsection{Qualitative Analysis}
\label{sec:qualitative}

We observe several consistent qualitative patterns: (1) the Progressive Injection Schedule produces notably smoother transitions, eliminating subtle color discontinuities visible with binary schedules; (2) Channel-Selective Perturbation preserves spatial structure more faithfully in object replacement tasks; (3) the combination produces cleaner edit boundaries, particularly for Add Object and Change Object tasks; (4) for Change Style edits, the Progressive Schedule alone provides measurable improvements by ensuring smoother feature transitions.

\subsection{Discussion}
\label{sec:discussion}

\begin{figure}[t]
\centering
\includegraphics[width=0.95\columnwidth]{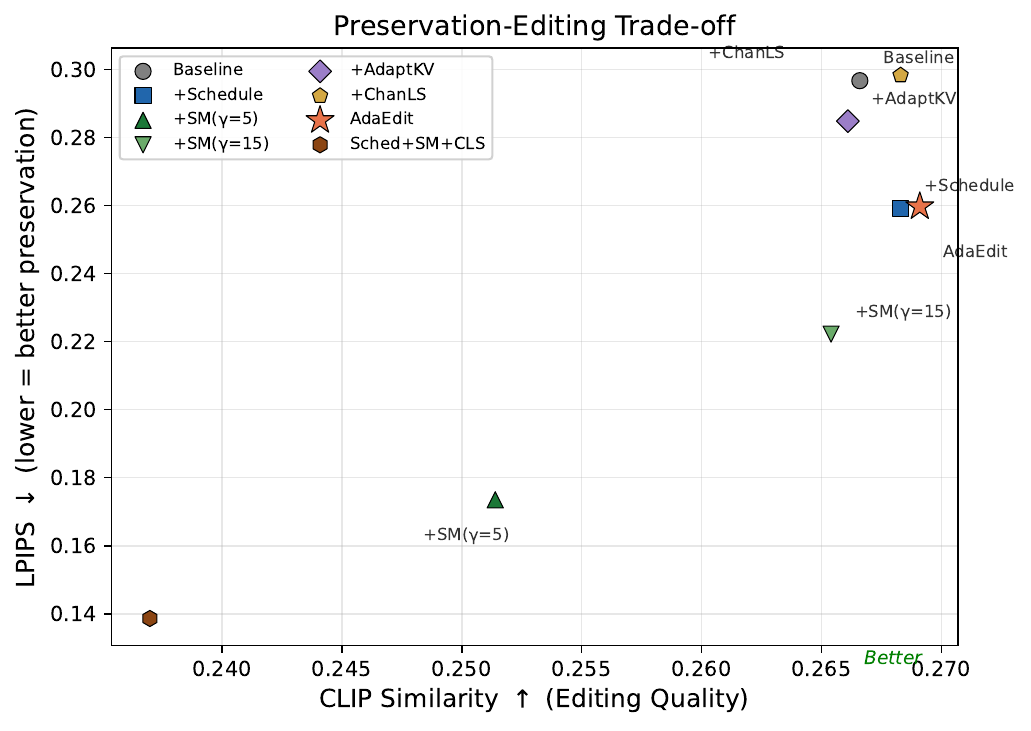}
\caption{Preservation-editing trade-off analysis. AdaEdit (orange star) achieves an optimal balance, while aggressive configurations like Soft Mask sacrifice editing quality for preservation.}
\label{fig:pareto}
\end{figure}

\textbf{Trade-off analysis.} The results reveal a fundamental trade-off in injection-based editing: stronger source feature preservation tends to come at the cost of editing flexibility. AdaEdit's key contribution is shifting this Pareto frontier: for a given level of CLIP similarity, AdaEdit achieves substantially better background preservation than the baseline. \cref{fig:pareto} visualizes this trade-off space across all evaluated configurations.

\textbf{Schedule choice.} Among the three schedule types, sigmoid performs best due to its ability to maintain near-full injection strength for the majority of the injection window before rapidly decaying.

\textbf{Temperature sensitivity.} The channel temperature $\tau$ controls the degree of channel differentiation. We find the method robust within $\tau \in [0.5, 2.0]$.

\textbf{Computational overhead.} AdaEdit introduces negligible overhead---total inference time remains within 1\% of the baseline.

\textbf{Limitations.} AdaEdit inherits the limitations of the underlying inversion-based editing paradigm: editing quality is bounded by inversion accuracy. The channel importance estimation relies on the assumption that distributional gap correlates with semantic importance. Our evaluation is conducted on FLUX-dev; generalization to other architectures requires further investigation. Edits requiring dramatic structural changes (\eg, Change Position) remain challenging.

\section{Conclusion}
\label{sec:conclusion}

We have presented AdaEdit, a training-free adaptive editing framework for flow-based image generation models. By identifying the temporal and channel heterogeneity of the injection demand---a fundamental but previously unaddressed aspect of inversion-based editing---we developed two complementary innovations: the Progressive Injection Schedule and Channel-Selective Latent Perturbation. The progressive schedule eliminates feature discontinuity artifacts caused by binary temporal cutoffs, while channel-selective perturbation focuses editing perturbation on semantically relevant channels while preserving structural ones. Extensive experiments on PIE-Bench demonstrate that AdaEdit achieves significant improvements in background preservation (8.7\% LPIPS reduction, 2.6\% SSIM improvement, 2.3\% PSNR improvement) with minimal impact on editing accuracy. As a plug-and-play framework compatible with multiple ODE solvers, AdaEdit provides a principled and practical improvement to the inversion-based editing paradigm. Future work will explore extending AdaEdit to video editing, adaptive temperature estimation, and integration with guidance-based editing methods.

{\small
\bibliographystyle{ieee_fullname}
\bibliography{references}
}

\end{document}